\documentclass[letterpaper, 10 pt, conference]{ieeeconf}  

\IEEEoverridecommandlockouts                              

\usepackage{dblfloatfix} 
\usepackage{graphicx}
\usepackage{url}
\usepackage{amsmath}
\usepackage{amsfonts}
\usepackage{flushend}

\title{\LARGE \bf
VAIR: Visuo-Acoustic Implicit Representations for Low-Cost, Multi-Modal Transparent Surface Reconstruction in Indoor Scenes
}

\author{Advaith V. Sethuraman$^{1*}$, Onur Bagoren$^{1*}$, Harikrishnan Seetharaman$^{1}$, \\ Dalton Richardson$^{1}$, Joseph Taylor$^{1}$, and Katherine A. Skinner$^{1}$
  \thanks{This work is supported by Amazon Consumer Robotics and a University of Michigan Robotics Department Fellowship.}
\thanks{$^{*}$denotes equal contribution.}
 \thanks{$^1$
 Department of Robotics, University of Michigan, Ann Arbor, MI 48109 USA. Corresponding author e-mail: {\tt\small advaiths@umich.edu}}
}

\begin{document}

\maketitle
\thispagestyle{empty}
\pagestyle{empty}

\begin{abstract}
Mobile robots operating indoors must be prepared to navigate challenging scenes that contain transparent surfaces. This paper proposes a novel method for the fusion of acoustic and visual sensing modalities through implicit neural representations to enable dense reconstruction of transparent surfaces in indoor scenes. We propose a novel model that leverages generative latent optimization to learn an implicit representation of indoor scenes consisting of transparent surfaces. We demonstrate that we can query the implicit representation to enable volumetric rendering in image space or 3D geometry reconstruction (point clouds or mesh) with transparent surface prediction. We evaluate our method's effectiveness qualitatively and quantitatively on a new dataset collected using a custom, low-cost sensing platform featuring RGB-D cameras and ultrasonic sensors. Our method exhibits significant improvement over state-of-the-art for transparent surface reconstruction. Website and Dataset: \url{https://umfieldrobotics.github.io/VAIR_site/} 

\end{abstract}


\section{Introduction}
Mobile robots navigating indoor spaces encounter various challenges for modern perception sensors, including highly reflective surfaces, poor lighting, and transparent objects. 
Popular sensors like RGB-D cameras and LiDAR cannot faithfully reconstruct transparent surfaces such as glass panes \cite{kuipers-glass, zhu2021rgbd}. When using 3D maps built using these sensors, robots will lack crucial geometric information about indoor spaces. This can lead to robots colliding with glass doors, floor-to-ceiling glass panes, and mirrored surfaces \cite{Ding2023AerodynamicEC}. 

Most prior work on robot perception for transparent objects and surfaces has focused on the manipulation of small transparent objects such as drinking glasses \cite{zhu2021rgbd, cleargrasp}. Although these methods work well in controlled table-top environments with fixed lighting, they are not well-equipped to handle transparent surface perception at the larger scale of indoor scenes. Prior approaches use techniques that work well when specularities and distortions are present on small glass objects \cite{IchnowskiAvigal2021DexNeRF}. However, these phenomena are rare and unreliable indicators of indoor glass structures, which are often planar. 

In this work, we explore the fusion of low-cost acoustic sensing modalities (i.e., ultrasonic sensors) with RGB-D imagery to enable reconstruction of transparent surfaces in indoor scenes.
\begin{figure}
\centering
\includegraphics[width=0.95\linewidth]{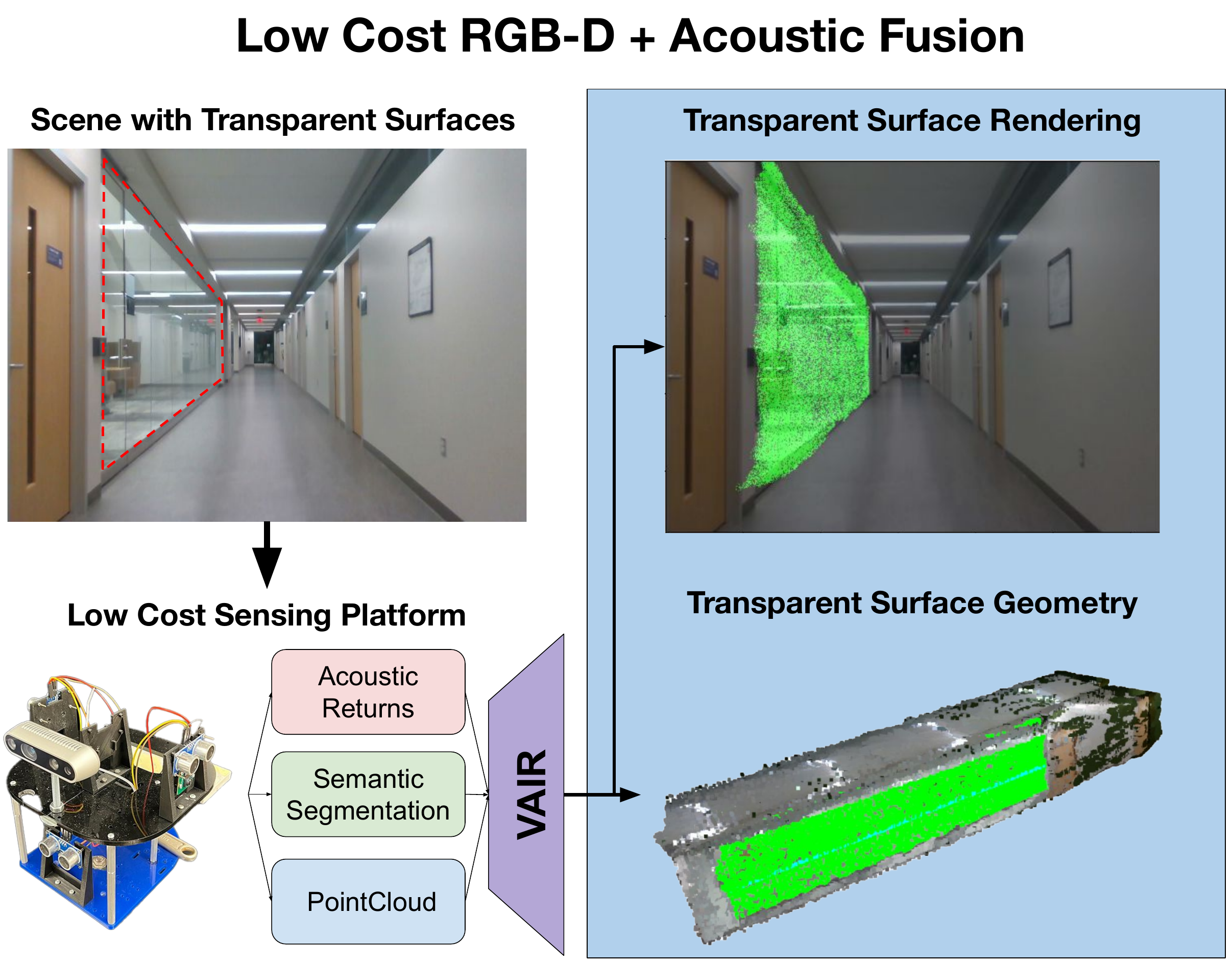}
\caption{We augment the mapping and reconstruction capabilities of mobile robots with low-cost acoustic sensors that can measure sparse returns from glass surfaces. Our framework, VAIR, fully reconstructs glass surfaces, producing useful 3D geometry for robotic systems. VAIR's transparent surface prediction is shown in green. \label{intro} \vspace{-6mm}}
\end{figure}
While many time-of-flight sensors that use electromagnetic radiation (e.g., structured light, LiDAR, etc.) fail to capture the geometry of transparent surfaces, we note that acoustic sensors are able to detect transparent surfaces reliably. However, acoustic sensors suffer from noise, interference, and, most importantly, measurement sparsity. To address this problem, we leverage the power of generative models to fuse the sparse readings of low-cost acoustic sensors with RGB-D information to reconstruct transparent surfaces in indoor scenes. 
Our proposed pipeline, illustrated in Fig. \ref{intro}, effectively fuses visual (RGB-D, semantic) and acoustic data to produce an implicit representation that we call \textbf{V}isuo-\textbf{A}coustic \textbf{I}mplicit \textbf{R}epresentation (VAIR). Concretely, we present the following contributions: 
\begin{itemize}
\item We develop a novel scene-conditioned generative model that produces an implicit representation of transparent surfaces in indoor scenes by fusing RGB-D point clouds, semantic segmentation predictions, and sparse acoustic sensor measurements. 
\item We develop a novel method for semantic-guided planar acoustic projection that initializes and guides our network training. 
\item We present a multi-modal dataset consisting of 3 real, indoor scenes with glass surfaces. Our dataset contains RGB-D images, acoustic sensor readings, camera poses, and ground truth geometry for glass surfaces. 
\end{itemize}

We demonstrate that VAIR exhibits a significant improvement over state-of-the-art 
in geometry prediction tasks, yielding more accurate 3D reconstructions of indoor scenes that feature transparent surfaces. 

\section{Related Works}

\subsection{Implicit Representations and 3D Generative Models}
Implicit representations have recently demonstrated impressive performance on scene reconstruction and novel view synthesis 
\cite{mildenhall2020nerf, li2023neuralangelo, mueller2022instant}. 
Recent work has explored leveraging implicit representations for robotic tasks including trajectory optimization \cite{nerf-nav}, simultaneous localization and mapping (SLAM) \cite{rosinol2022nerf,isaacson2023loner}, and semantic scene understanding \cite{Zhi:etal:ICCV2021}. 
Most relevant to this work, prior work has explored implicit representations as a natural framework for multi-sensor fusion \cite{Zhu2023, schmid2024virusnerf}. Zhu et al. present the Multi-Modal Radiance Field, which fuses infrared (IR), RGB images, and point clouds to produce a representation of the scene \cite{Zhu2023}. VIRUS-NeRF demonstrates fusion of low-cost sensing modalities, including IR, ultrasonic sensing, and RGB images to produce occupancy grids of a robot's surroundings \cite{schmid2024virusnerf}.

The marriage of generative models with implicit representations has led to improved scene and view synthesis capabilities \cite{pmlr-v139-kosiorek21a, liu2021editing, liu2023zero1to3, Park_2019_CVPR}. An appealing property of generative models is the ability to \textit{extrapolate} well from sparse measurements. For example, DeepSDF presents a method to generate signed distance functions (SDFs) from sparse point cloud measurements \cite{Park_2019_CVPR}. Liu et al. use user-supplied scribbles in image space to edit the colors and geometries of implicit representations of objects \cite{liu2021editing}. 


In this work, we leverage recent advances in implicit representations and generative models to enable multi-sensor fusion of RGB-D imagery and sparse acoustic data from ultrasonic sensors to produce a dense 3D reconstruction of transparent surfaces for robotics applications. 

\subsection{Perception for Transparent Objects and Surfaces}

Transparent objects are commonplace in indoor scenes and 
pose challenges for robot perception algorithms. Prior work has focused on the perception tasks of depth completion, semantic segmentation, and pose estimation for transparent objects \cite{zhu2021rgbd, IchnowskiAvigal2021DexNeRF, chad_light_field, fang2022transcg, NEURIPS2022_8d162f48,jiang2023robotic, NerfREN, graspnerf,gw-depth}. Our work primarily focuses on the fusion of acoustic, visual, and semantic information for producing an \textit{implicit representation} of the scene to learn to model the 3D geometry of a scene with reconstruction of transparent surfaces 

Recent work has focused on implicit representations for modeling transparent objects in table-top scenes for robot manipulation. Zhu et al. leverage implicit functions of ray/voxel pairs to complete missing depth measurements from an RGB-D camera to reconstruct transparent objects such as beakers and cups \cite{zhu2021rgbd}. SAID-NeRF leverages semantic information to improve reconstruction quality of transparent objects \cite{ummadisingu2024saidnerf}. These methods benefit from the large number of available table-top datasets with ground truth depth or pose estimation and multi-view perspectives, which can be easily collected for table-top scenes.

Other work has considered multi-modal sensing to aid in perception for transparent objects. LIT leverages light field cameras to fit CAD models to transparent objects for pose estimation \cite{chad_light_field}. Kim et al. propose the use of multi-spectral imaging for transparent object detection and pose estimation \cite{transpose}. Most similar to our proposed sensor configuration, Zhang et al. leverage sparse acoustic sensors (ultrasonic) along with RGB-D images in a probabilistic formulation to augment a KinectFusion pipeline for reconstructing transparent objects and surfaces in the scene \cite{manocha}. More recently, Weerakoon et al. propose TOPGN, a LIDAR-based method that is able to detect transparent surfaces in the scene. Though effective, we choose to focus on low-cost modalities such as RGB-D cameras and acoustic sensors. 
Still, Zhang et al. assume collection of ultrasonic returns both vertically and horizontally on transparent surfaces with a handheld sensor rig. Having sufficient coverage of the transparent surface with ultrasonic returns allows their method to fit geometry effectively. Our work instead focuses on perception of transparent surfaces for mobile robot navigation where we do not have controlled lighting and we often do not capture such controlled multi-view perspectives of the scene. 

Most similar to our work, NeRFRen leverages implicit representations for sensing and reconstructing transparent and reflective surfaces~\cite{NerfREN}. Citing the shortcoming of standard NeRF methods in modeling reflective and transmitted objects for novel view synthesis, NeRFRen decomposes the problem into two different representations: transmitted and reflected maps. The view synthesis is then done as a weighted addition between the two rendered images. The method introduces geometric and semantic priors for guiding the learning of the reflective and transmitted maps. Prior work on semantic segmentation of transparent objects and surfaces in indoor scenes has demonstrated promising results on real-world datasets~\cite{NEURIPS2022_8d162f48}. Our proposed method also leverages semantic information. However, our framework allows for multi-sensor fusion of low-cost sensing modalities to enable dense 3D reonstruction of transparent surfaces. We compare the performance of VAIR to NeRFReN and demonstrate that the addition of the complementary acoustic sensing modality significantly improves transparent surface reconstruction.

\section{Method}
\begin{figure*}[]
\vspace{4mm}
\centering
\includegraphics[width=0.98\linewidth]{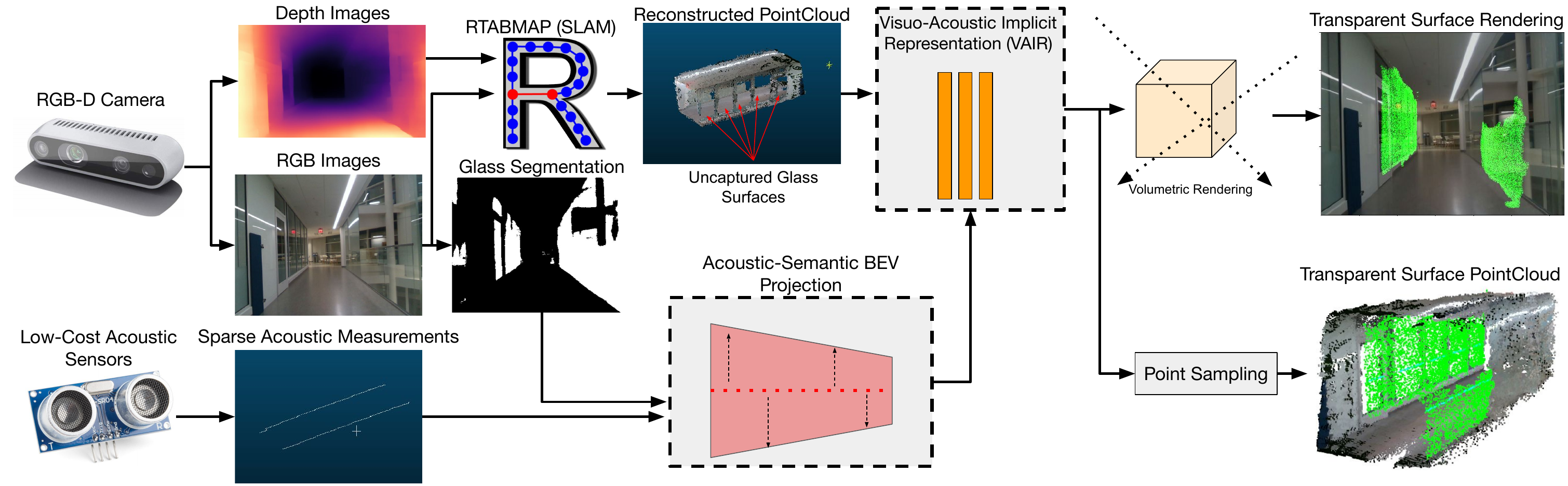}
\caption{VAIR is designed to integrate into existing robotic SLAM pipelines. The fusion of sparse acoustic sensor measurements and RGB-D imagery allows us to sense and reconstruct transparent surfaces. VAIR exploits semantic information from RGB images to further inform prediction and reconstruction of transparent surfaces and to learn an implicit representation of the scene. This representation can be queried for downstream robotic tasks. \label{overview}}
\end{figure*}

Figure \ref{overview} provides an overview of our proposed framework. Our inputs are RGB-D images and sparse acoustic measurements from low-cost ultrasonic sensors. The RGB-D images are input to RTABMap to produce a 3D point cloud reconstruction of the environment \cite{labbe2019rtab}. Note that RTABMap is not able to reconstruct transparent surfaces with high fidelity. The RGB images are also input to an off-the-shelf semantic segmentation network to produce segmentation masks for transparent surfaces \cite{neurips2022:gsds2022}. Given the sparse acoustic measurements and semantic segmentation, we propose an acoustic-semantic planar projection (ASPP) to align the ultrasonic measurements with the transparent surface segmentations. Finally, the reconstructed point cloud and ASPP are used by VAIR to generate an implicit representation of transparent surfaces in the scene. This representation can be queried to render the location of transparent surfaces in image space or sampled to produce 3D geometry such as a point cloud.

\subsection{Acoustic-Semantic Planar Projection (ASPP)}
Our work proposes to leverage multi-sensor fusion of RGB-D data and sparse ultrasonic readings for transparent surface reconstruction. The benefit of acoustic data is that it can provide information about the existence of transparent surfaces. However, the acoustic measurements are sparse and there is an inherent ambiguity in the exact location of the return. Following Zhang et. al., we assume that the acoustic return comes from $0^{\circ}$ azimuth and $0^{\circ}$ elevation \cite{manocha}. We find that this assumption works well for the environments of interest in this work. 
We propose a pre-processing step that leverages semantic priors to project the sparse acoustic measurements into 3D space. Note that although there exist state-of-the-art glass semantic segmentation models \cite{neurips2022:gsds2022}, they perform segmentation in \textit{pixel space}, and producing useful 3D geometry is still an open problem.  
\begin{figure}[b!]
\includegraphics[width=0.95\linewidth]{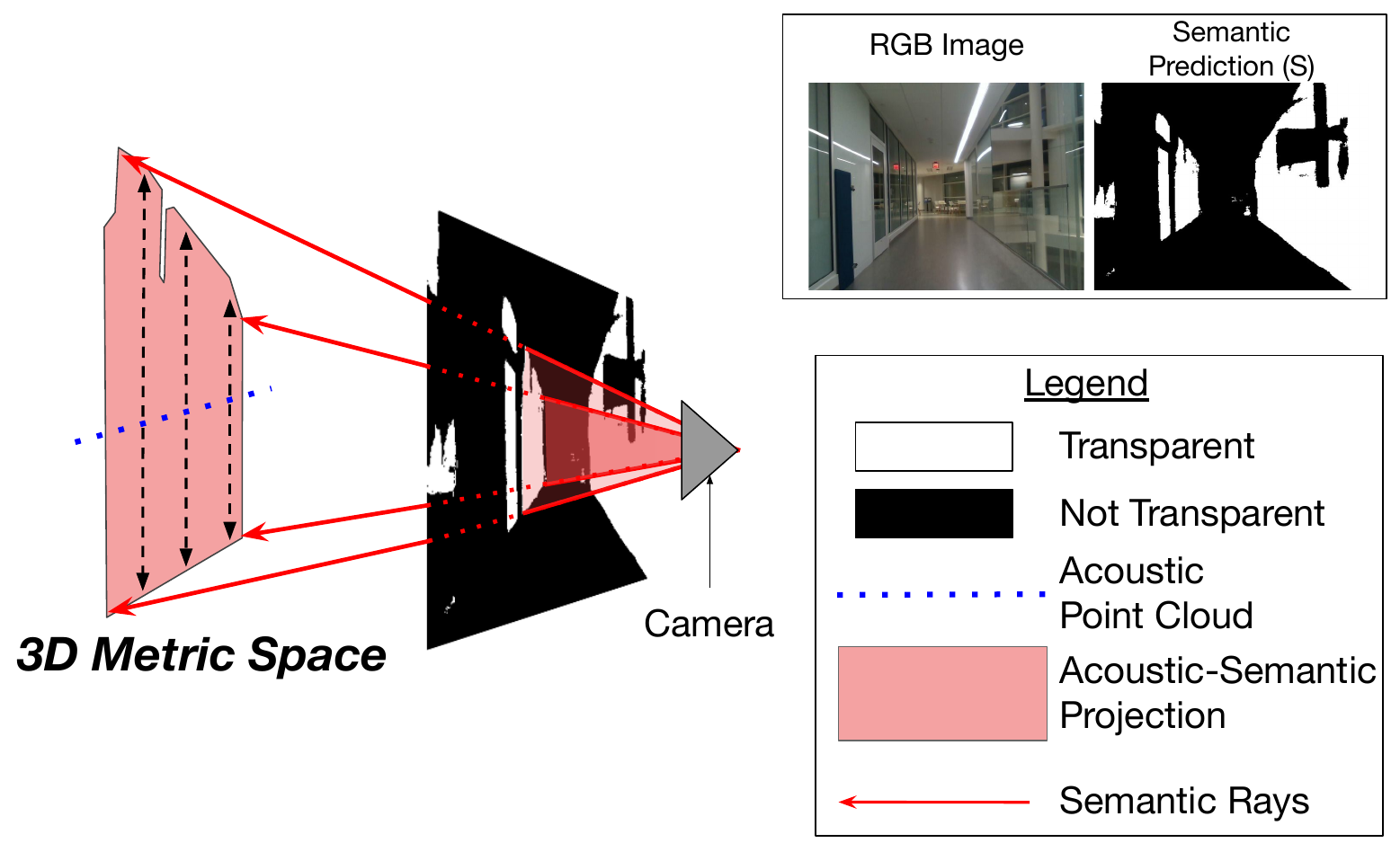}
\caption{Our Acoustic-Semantic Planar Projection (ASPP). We project rays through pixels predicted as transparent surface pixels, as specified by semantic segmentation, to further inform the extents of transparent surfaces in the scene. The projection of acoustic points onto the semantic rays is provided as input to VAIR. \label{ASP}}
\end{figure}
\begin{figure*}[ht!]
\vspace{4mm}
\centering
\includegraphics[width=0.95\linewidth]{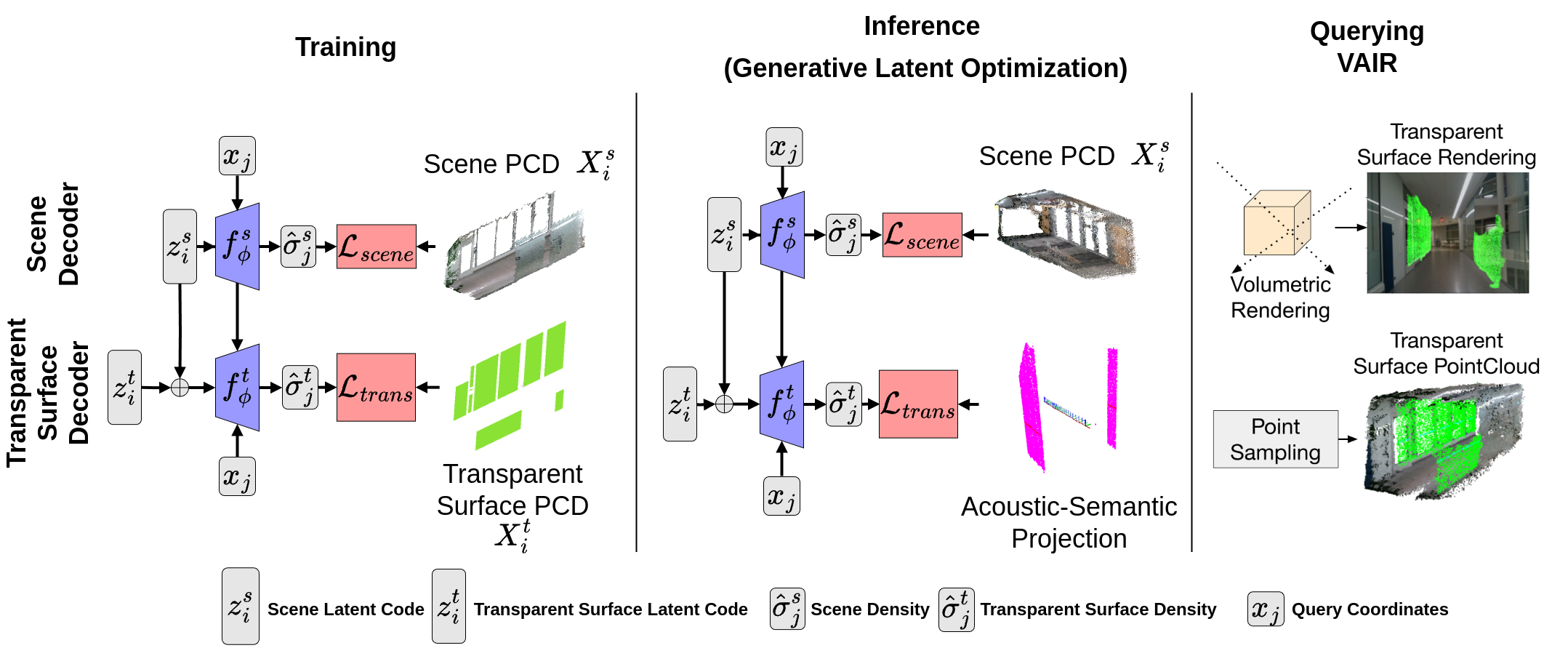}
\caption{Overview of VAIR. During training, VAIR takes as input latent codes for the scene and transparent surfaces. The respective decoders map the latent codes to density values $\hat{\sigma}^s_j$ and $\hat{\sigma}^t_j$ supervised by the scene point cloud $X^s_i$ and transparent surface point cloud $X^t_i$. During test time, we perform a generative latent optimization with randomly initialized latent codes. The latent codes are passed to decoders and losses are computed on density values $\hat{\sigma}^s_j$ and $\hat{\sigma}^t_j$ against scene geometry, semantic information, and sparse acoustic measurements in the form of the ASPP. VAIR is able to predict an implicit density field that completes the scene after finding latent codes that best explain the partial scene geometry. \label{method}}
\vspace{-6mm}
\end{figure*}

Figure~\ref{ASP} illustrates our proposed method for ASPP. 
Consider an acoustic point cloud ($APC$) and semantic segmentation prediction $S$. 
Let $\mathcal{G} = \{(u,v), S[u,v] = 1\}$ be the set of pixels that are predicted as transparent surfaces in the image. We find the ray $r_{u,v}$ that passes through every pixel $g \in \mathcal{G}$ using the extrinsic and intrinsic matrix of the camera. Then, for every point in the APC, we consider a radius of $\epsilon$ in the xy-plane. This ``pillar" is extended in the z-axis using the min/max z-values of the semantic rays that intersect it. This provides an initial estimate of the transparent surface informed by both the acoustic sensing and semantic predictions.

\subsection{VAIR Architecture \label{vair}}
Figure~\ref{method} illustrates the training and inference architecture for the VAIR network. VAIR takes input latent codes and learns to output an implicit representation of opaque and transparent surfaces in the scene. We propose a scene-conditioned generative model to learn the distribution of indoor scenes and transparent surfaces within those scenes. Since a given indoor scene can have any variety of configurations of transparent surfaces (open/closed doors, glass panes, or windows), we formulate our model to disentangle the representations for non-transparent and transparent surfaces within the scene. VAIR consists of a scene decoder and a transparent surface decoder, each with decoder-specific latent codes. VAIR is trained on synthetic data, detailed in Section~\ref{training}. 
During inference, we perform a test-time optimization to find optimal latent codes that best explain our sensor measurements.

VAIR is formulated as a generative latent optimization framework. Generative latent optimization (GLO) aims to learn representations of data using \textit{decoder-only} networks \cite{GLO} \cite{Park_2019_CVPR}. GLO has a natural ability to handle partial inputs, such as sparse sensor readings. With VAIR, we present a novel formulation of GLO for modeling transparent surfaces within indoor scenes and their corresponding density fields.

First, consider the general case of a dataset $X_i$ of $N$ 3D geometries and their corresponding density fields: 

\begin{equation}
X_i = \{(x_j, \sigma_j): \sigma_j = DF_i(x_j)\}
\end{equation}

\noindent where $x_j$ is a 3D query point, $\sigma_j$ is the density value, and $DF_i(\cdot)$ is a density field function that maps query points in the $i^{th}$ geometry to corresponding density values. 
In this work, we enforce points on surfaces to have density value $\sigma_{max}$ and free space to have density value 0. This is done to ensure natural integration with popular volume-rendered implicit representations such as NeRF \cite{mildenhall2020nerf}. For each data point $X_i$, we assign latent code $z_i \sim \mathcal{N}(0, \sigma_z^2)$ that will be jointly optimized along with decoder $f_{\theta}$. 





During training time, for the $i^{th}$ data point, we optimize the following objective function: 

\begin{equation}
\mathcal{L}_{scene} = \sum_{(x_j, \sigma_j) \in X_i^s} ||f^s_{\theta}(z_i^s, x_j) - \sigma_j||_2^2 + \frac{1}{\sigma_z^2}||z_i^s||_2^2
\end{equation}



Our transparent surface decoder $f^{t}_{\phi}$ learns the distribution of \textit{only} transparent surfaces in the scenes. Since the location and extents of transparent surfaces can be informed by the surrounding opaque geometry of the scene (i.e. frames of a window, opaque walls around a glass pane), we condition $f^t_{\phi}$ with scene latent code $z^s_i$. 
We train the transparent surface decoder and transparent surface latent codes $z^t = \{z_i^t\}_{i=1}^N$ using the following objective function: 

\begin{equation}
\mathcal{L}_{trans} = \sum_{(x_j, \sigma_j) \in X_i^t} ||f^t_{\phi}(z^t_i \oplus z_i^s, x_j) - \sigma_j||_2^2 + \frac{1}{\sigma_z^2}||z_i^t||_2^2
\end{equation}

\noindent where $\oplus$ denotes concatenation. During training, we update decoder weights ($\theta, \phi$) and latent codes ($z^s, z^t$). Our final loss function becomes: 

\begin{equation}
\mathcal{L}_{train} = \mathcal{L}_{scene} + \mathcal{L}_{trans}
\end{equation}


During inference, we freeze decoder weights $\theta, \phi$ but update latent codes $z^s, z^t$ to minimize the following loss function: 
\begin{equation}
\mathcal{L}_{inf} = \mathcal{L}_{scene} + \mathcal{L}_{trans}
\label{inf}
\end{equation}

\noindent where $\mathcal{L}_{trans}$ is calculated using $APC \cup P_{ASP}$. If we consider $z^{s*}, z^{t*}$ to be minimizers of Eq. (\ref{inf}), the final transparent surface density field output is given by: $f_{\phi}^t(z^{s*} \oplus z^{t*}, \cdot)$. Similar to NeRFs, $f_{\phi}^t(z^{s*} \oplus z^{t*}, \cdot)$ can be queried at 3D coordinates to produce a density value. We are able to volumetrically render this density field in the camera frame or transform it into a mesh using marching cubes. 

\section{Experiments and Results}

\begin{table*}[hbtp!]
\centering
\vspace{4mm}
\caption{Quantitative results for 3D reconstruction of transparent surfaces in the FRB Sequences dataset. We report masked and unmasked voxelized intersection-over-union (IOU$_m$ and IOU$_{un}$) for the \textbf{GLASS} class. Chamfer Distance is CD-$\ell_1$ $\times$ 10$^3$. \label{quant}}
\scalebox{0.85}{
\begin{tabular}{l|ccc|ccc|ccc|ccc}
               & \multicolumn{3}{c|}{Hallway}                  & \multicolumn{3}{c|}{Balcony}                  & \multicolumn{3}{c|}{Conf}                      & \multicolumn{3}{c}{Average}                    \\ \hline
Method         & IOU$_m$ $\uparrow$      & IOU$_{un}$  $\uparrow$       & CD-$\ell_1$ $\downarrow$                   & IOU$_m$  $\uparrow$        & IOU$_{un}$  $\uparrow$       & CD-$\ell_1$ $\downarrow$                      & IOU$_m$ $\uparrow$         & IOU$_{un}$  $\uparrow$       & CD-$\ell_1$ $\downarrow$                      & \multicolumn{1}{c}{IOU$_m$ $\uparrow$}  & \multicolumn{1}{c}{IOU$_{un}$ $\uparrow$}  & \multicolumn{1}{c}{CD-$\ell_1$ $\downarrow$}             \\ \hline
Depth Anything v2 \cite{depth_anything_v2} &0.00&	0.07&	1,144.00&	0.11&	0.06&	496.75&	0.32&	0.12&	219.71&	0.14&	0.08&	620.15         \\
NeRFRen \cite{NerfREN}    & 0.13&	0.10	&525.64&	0.22&	0.17	&323.76	&0.10&	0.13&	560.47&	0.15&	0.13&	469.96                                 \\
VAIR (ours)     &  \textbf{0.89}&	\textbf{0.33}&	\textbf{24.52}&	\textbf{0.79}&	\textbf{0.38}&	\textbf{37.25}&	\textbf{0.39}&	\textbf{0.39}&	\textbf{73.98}&	\textbf{0.69}&	\textbf{0.37}&	\textbf{45.25}
\end{tabular}
}
\end{table*}

\subsection{Implementation Details}

The scene decoder $f^s_{\theta}$ and transparent object decoder $f^t_{\phi}$ follow a 3D convolutional decoder architecture.  $f^s_{\theta}$ takes as input latent vectors of size 256, whereas $f^t_{\phi}$ takes in latent vectors of size 8. We found this distinction to be effective because full scenes require more information to encode than the transparent surfaces of interest. Both $z^s$ and $z^t$ are initialized with zero-mean Gaussian noise with standard deviation $\sigma_z$ = 0.01. Both networks output a voxel-grid of size $(128, 128, 128)$, which is then trilinearly sampled at the desired points to produce density values $\sigma_j^s$ and $\sigma_j^t$. We found experimentally that $\sigma_{max} = 100$ worked well during training. We train our model for 250 epochs on 4 NVIDIA A100 GPUs with 84GB of VRAM with a learning rate of 1e-3. We perform SGD (25 iterations) to optimize for latent codes $z^s, z^t$ during inference using a learning rate of 8e-3. 

\subsection{Synthetic Training Data Generation\label{training}}
VAIR is trained using synthetic data that features paired point clouds: the transparent surface point cloud, $X^t_i$, which provides ground truth geometry for transparent surfaces, and the opaque scene point cloud, $X^s_i$, which provides ground truth geometry for the opaque surfaces in the scene. 
To generate this synthetic dataset, we leverage available indoor datasets to produce large amounts of point cloud training data of indoor scenes~\cite{Matterport3D,xiazamirhe2018gibsonenv,replica19arxiv}. 
 First, we perform a sliding-window non-overlapping crop of all meshes with a fixed crop size of $(3m, 3m, 4m)$. Then, we sample 500k points on the cropped mesh to generate a point cloud. We model three types of common indoor transparent surfaces: floor-to-ceiling glass panes, half-panes, and windows. 
Depending on the type of synthetic transparent surface, we choose cutout centers on the walls using normals of the mesh. We cut out points in the point cloud in the shape of a transparent surface and add the cutout points to $X^t_i$. 
The remaining points in the scene are added to $X^s_i$. 
We randomize the number of transparent surfaces, their dimensions, and their locations within the scene. Our synthetic dataset consists of 25,406 total point cloud pairs of indoor scenes with transparent surfaces. 

\subsection{Ford Robotics Building (FRB) Dataset}
\begin{figure}
    \centering
    \includegraphics[width=0.7\linewidth]{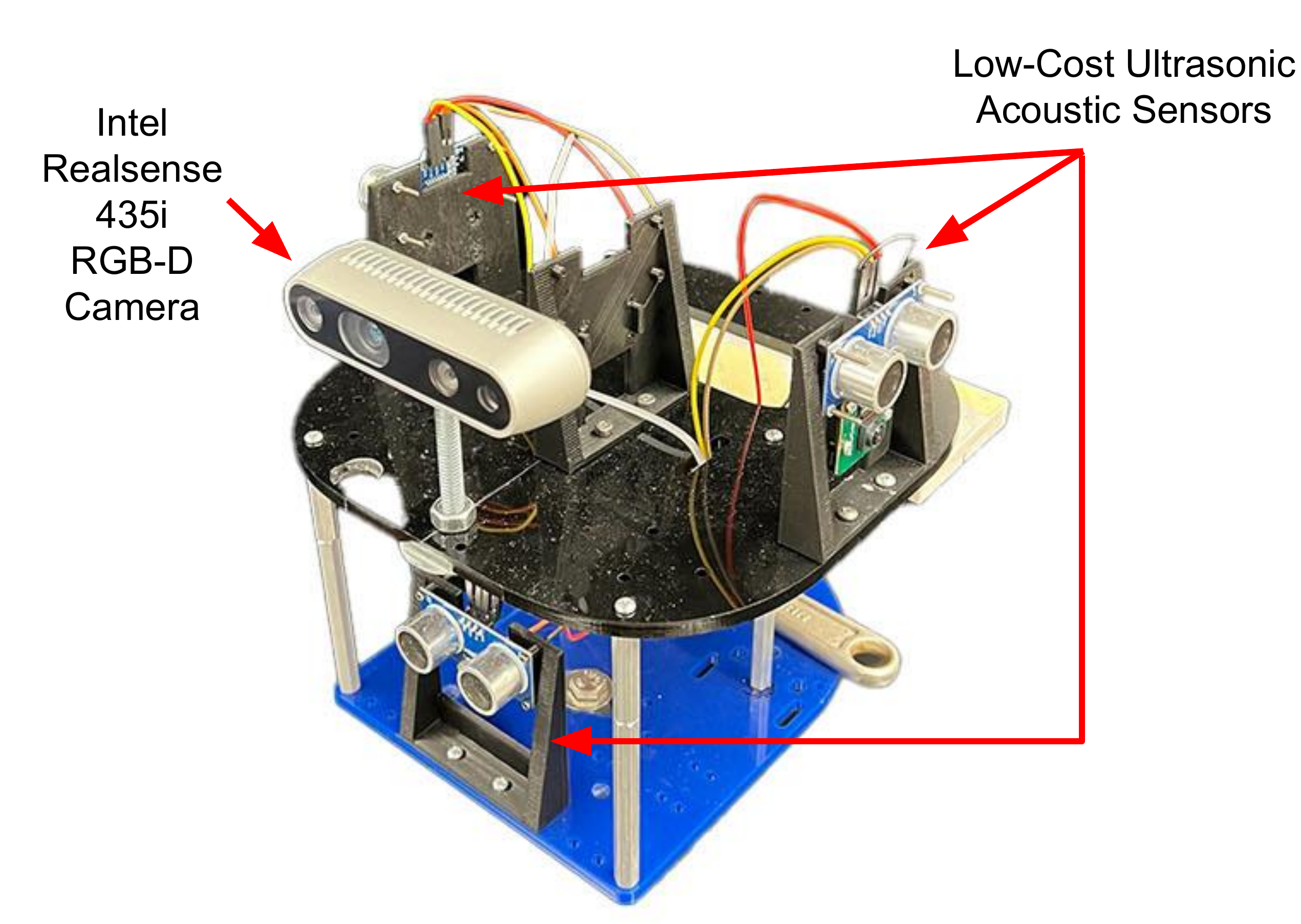}
    \caption{Our custom sensing platform consists of an array of low-cost acoustic sensors, an Arduino, and a forward-facing Intel Realsense 435i RGB-D camera. 
    \label{sensor_rig}}
    \vspace{-4mm}
    
\end{figure}

There is limited availability of open-source datasets that have acoustic sensor measurements and labeled transparent surfaces. To address this, we collect a real-world dataset with ground truth for transparent surface geometry. Figure~\ref{sensor_rig} shows our sensing platform, which consists of three low-cost ultrasonic acoustic sensors and an Intel Realsense 435i RGB-D camera. 
We place the acoustic sensors such that they are pointing at perpendicular and opposite directions to geometrically limit multi-path interference, which is a common phenomena for acoustic sensors that trigger a return whenever a signal at a specific frequency is returned.
We model the acoustic sensor as a simple point range measurement. 
Integration of a higher fidelity model of the view frustum is left for future work~\cite{schmid2024virusnerf, qadri2022neural}. 

To transform acoustic sensor measurements into the RGB-D camera frame, we manually measure the extrinsics. An Arduino Uno is used to relay the acoustic sensor information to a data collection computer via serial/UART. The RGB-D camera captures images at 30Hz, whereas the acoustic sensors are timed to emit pings at approximately 10Hz to avoid multi-path interference. Our sensing platform is integrated within a ROS framework. 
We mount the sensing platform on a cart, which emulates a mobile robot and can be easily navigated through indoor scenes. 

We capture three indoor scenes -- \textit{Hallway}, \textit{Conference Room}, and \textit{Balcony} -- which contain a variety of transparent surfaces. 
To produce ground truth geometry for evaluation, we use the RGB images and 3D point cloud output by RTABMap as references to manually place planes at the locations of transparent surfaces. Our dataset will be made publicly available for future research in visuo-acoustic sensor fusion for transparent surface perception.

\begin{figure*}[h!]
\vspace{1mm}
   \centering
    \includegraphics[width=0.89\linewidth]{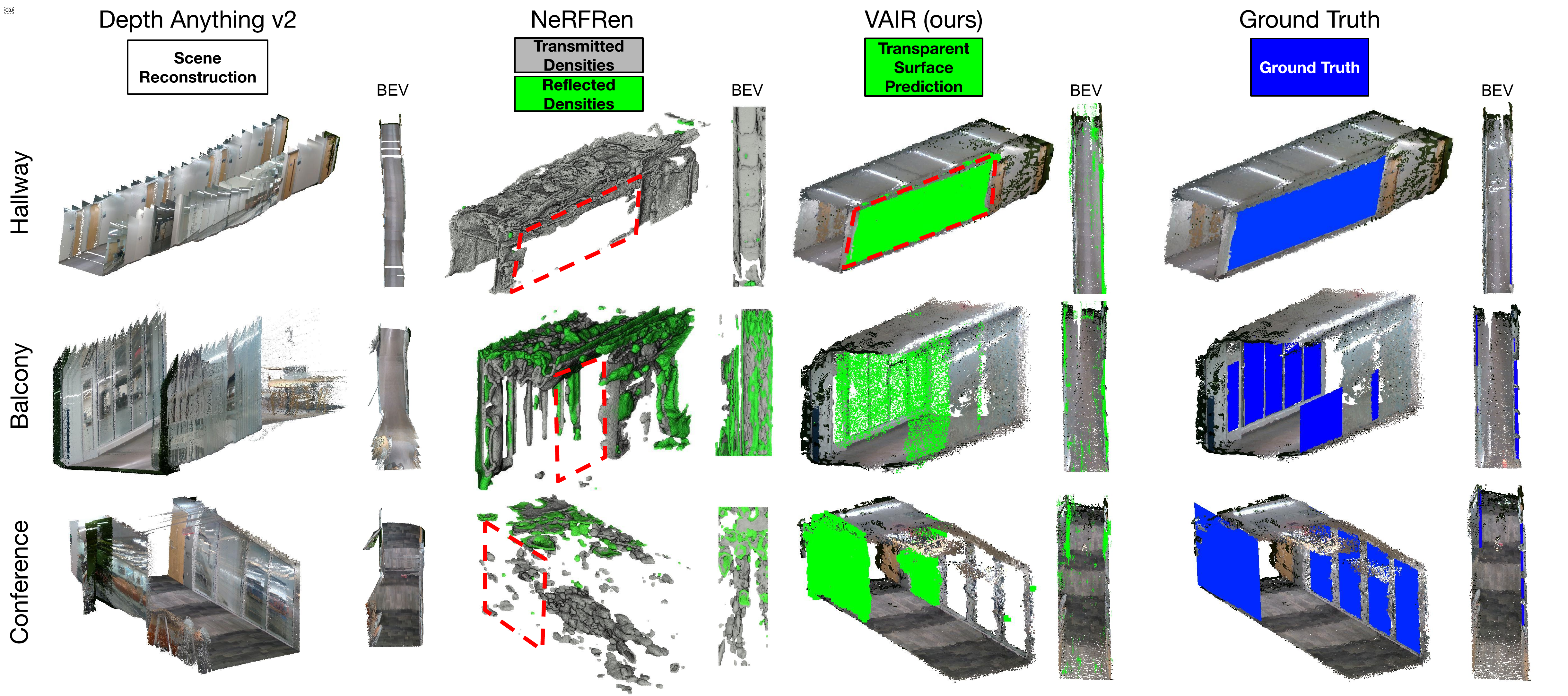}
    \caption{Point cloud visualization from Depth Anything v2, NeRFRen, and VAIR. Note that Depth Anything v2 does not produce a separate transparent surface prediction, but still is able to predict reasonable depths on transparent surfaces from the RGB image. VAIR can produce a point cloud by sampling points in 3D space and thresholding based on the density value. We use a density threshold of $\sigma = 85$ for all scenes. We also include Birds-Eye-View (BEV) of the reconstructions. NeRFRen's opaque surfaces are visualized in gray and transparent surfaces are visualized in green since marching cubes does not return colors. A different shader was used for better visibility.\label{pcd_results}}
    \vspace{-4mm}
\end{figure*}
\subsection{Transparent Surface Reconstruction}
We evaluate performance qualitatively and quantitatively with comparison between our method, VAIR, NeRFRen~\cite{NerfREN}, and Depth Anything v2~\cite{depth_anything_v2} integrated with RTABMap on the FRB dataset. To generate reconstructions for VAIR, we sample 500k points randomly within the bounds of the implicit function output. Then, we collapse the points onto the density field predicted by VAIR. We voxelize this point cloud using a voxel grid of resolution $64^3$. To produce point clouds for NeRFRen, we use marching cubes to first produce a mesh. We uniformly sample 500k points on the surface of the mesh and voxelize the point cloud at a resolution of $64^3$. To produce point clouds for Depth Anything v2, we use the metric depth prediction network to produce depth images from RGB images and input the predicted depth and RGB images to RTABMap to produce a 3D reconstruction. 
To produce the ground truth voxel grid, we sample points on the faces of the ground truth mesh and perform the same voxelization procedure. 

Following \cite{Azinovic_2022_CVPR}, we use voxelized intersection-over-union (IOU) and chamfer distance (CD-$\ell_1$) as our primary evaluation metrics. We use unmasked and masked IOU to evaluate reconstruction performance for both the full scene and for scene regions containing only transparent surfaces, respectively. Given a prediction of the transparent voxel grid $\hat{Y}$ and the ground truth transparent voxel grid $Y$, masked IOU is defined as  $IOU_{m} = \frac{(\hat{Y} \cap Y) \cap Y}{\hat{Y} \cup Y}$. Unmasked IOU is defined traditionally as $IOU_{un} = \frac{\hat{Y} \cap Y}{\hat{Y} \cup Y}$. We compute chamfer distance using the L1 norm on the entire scene (transparent and opaque surfaces). 

Table~\ref{quant} provides quantitative results. Note that NeRFRen is unable to faithfully represent the transparent surfaces across all scenes. Depth Anything v2 is able to predict plausible depth values on transparent surfaces, but they are not accurate to the ground truth, yielding low IOU. Finally, VAIR outperforms all baselines.  

Figure \ref{pcd_results} shows qualitative results of scene reconstruction. NeRFRen suffers from inaccurate transparent surface estimation. 
Specifically, in the \textit{Hallway} scene, we show that while NeRFRen is unable to produce sufficient density on the long glass pane, VAIR, informed by the acoustic point clouds, is able to produce a faithful representation of glass. 
In another scene, \textit{Balcony}, we show that in the absence of distinctive enough visual features for NeRFRen to use for transparent surface reconstruction the detection critically fails.
Depth Anything v2 is able to predict metric depth images given an input RGB image. In the \textit{Hallway} scene, Depth Anything v2 predicts the transparent surface as curved outwards. VAIR, on the other hand, guided by the multi-modal sensing capabilities, faithfully detects and reconstructs the glass surfaces.


\subsection{Timing Studies}
Training VAIR 
took approximately 48 hours on 4 NVIDIA A100 GPUs. Training NeRFRen takes 8 hours per scene with hyperparameters suggested in \cite{NerfREN} on a single NVIDIA A100 GPU. 
For timing studies, we assume RTABMap is running concurrently and report numbers that are averaged across all scenes. The total inference time for VAIR includes the glass segmentation network, ASPP creation, generative latent optimization, and point cloud sampling. VAIR takes 2.56 seconds for inference (0.39 Hz average). The total inference time for Depth Anything v2 includes only the forward pass for the metric depth prediction network. Depth Anything v2 takes 0.32 seconds for inference (3.13 Hz average). The total inference time for NeRFRen is 2.73 seconds including performing marching cubes and sampling a point cloud.

\subsection{Ablation Study}
Table~\ref{abl} shows an ablation study to quantify transparent surface reconstruction performance of VAIR compared to just using the ASPP or RTABMap directly. 
We find that although the ASPP works well, adding the generative capabilities of VAIR further enhances the performance by \textbf{18.33\%}, which indicates the benefit of the generative capabilities of VAIR for producing high-quality transparent surface reconstructions. RTABMap, which relies on RGB-D data only, fails to reconstruct transparent surfaces.

\begin{table}[]
\centering
\caption{Ablation study (IOU$_{m}$). \label{abl}}
\vspace{-4mm}
\begin{tabular}{lccc|c}
\hline
   Method        & Hallway & Balcony & Conf. Room  & Average   \\ \hline
RTABMap Only & 0.00  & 0.01 & 0.02 & 0.01\\
ASPP Only   & 0.76             & 0.61             & 0.43      & 0.60      \\
VAIR (ours) & \textbf{0.84}             & \textbf{0.76}             & \textbf{0.53}      & \textbf{0.71}     
\end{tabular}

\end{table}

\section{Conclusion and Future Work}
We present VAIR, a scene-conditioned, multi-modal generative model that is able to reconstruct geometry of indoor scenes featuring transparent surfaces. We demonstrate that VAIR significantly improves 3D reconstruction of transparent surfaces. 
One assumption we make for ASPP is that the transparent surfaces are planar, which is applicable for many glass surfaces in indoor scenes. For future work, we will 
extend our generative model to other transparent objects like boxes, display cases, and tables. To enable practical deployment, we will also investigate optimization techniques to reduce our fine-tuning and inference times for operation on low-cost mobile robotics platforms \cite{10342067}.



\bibliographystyle{ieeetr}
\bibliography{biblio}

\end{document}